\documentclass{article}

\usepackage{arxiv}

\usepackage[utf8]{inputenc} 
\usepackage[T1]{fontenc}    
\usepackage{graphicx}
\usepackage{caption}
\usepackage{subcaption}
\usepackage{tabularx}
\usepackage{array}
\usepackage{amsmath,amssymb,amsfonts}
\usepackage{nccmath}
\DeclareMathOperator*{\argmax}{arg\,max}

\usepackage{algorithm}
\usepackage{algorithmic}
\usepackage{multirow}
\usepackage{textcomp}
\usepackage{xcolor}
\usepackage{hyperref}
\hypersetup{
    colorlinks=true,
    linkcolor=blue,
    filecolor=magenta,      
    urlcolor=cyan,
}
\usepackage{cite}
\usepackage{appendix}

\title{Multi-robot Cooperative Object Transportation using Decentralized Deep Reinforcement Learning }

\author{
    Lin~Zhang\thanks{Intelligent Robotics and Autonomous System Lab, University of Cincinnati, Cincinnati, USA. Correspondence to: Lin Zhang<lin.zhang@uc.edu>} \\
    Department of Aerospace Engineering and Engineering Mechanics\\
    University of Cincinnati\\
    Cincinnati, OH 45221 \\
    \texttt{lin.zhang@uc.edu} \\
    \AND
    Hao~Xiong \\
    School of Mechanical Engineering and Automation \\
    Harbin Institute of Technology\ \
    Shenzhen, China 518055 \\
    \texttt{xionghao@hit.edu.cn} \\
    \AND
    Ou~Ma \\
    Department of Aerospace Engineering and Engineering Mechanics\\
    University of Cincinnati\\
    Cincinnati, OH 45221 \\
    \texttt{maou@ucmail.uc.edu} \\
    \AND
    Zhaokui~Wang \\
    School of Aerospace Engineering \\
    Tsinghua University \\
    Beijing, China 100084 \\
    \texttt{wangzk@tsinghua.edu.cn} \\
}

\begin{document}
\maketitle

\begin{abstract}
Object transportation could be a challenging problem for a single robot due to the oversize and/or overweight issues. A multi-robot system can take the advantage of increased driving power and more flexible configuration to solve such a problem. However, increased number of individuals also changed the dynamics of the system which makes control of a multi-robot system more complicated. Even worse, if the whole system is sitting on a centralized decision making unit, the data flow could be easily overloaded due to the upscaling of the system. In this research, we propose a decentralized control scheme on a multi-robot system with each individual equipped with a deep Q-network (DQN) controller to perform an oversized object transportation task. DQN is a deep reinforcement learning algorithm thus does not require the knowledge of system dynamics, instead, it enables the robots to learn appropriate control strategies through trial-and-error style interactions within the task environment. Since analogous controllers are distributed on the individuals, the computational bottleneck is avoided systematically. We demonstrate such a system in a scenario of carrying an oversized rod through a doorway by a two-robot team. The presented multi-robot system learns abstract features of the task and cooperative behaviors are observed. The decentralized DQN-style controller is showing strong robustness against uncertainties. In addition, We propose a universal metric to assess the cooperation quantitatively.
\end{abstract}

\section{Introduction}
In the world of humans, complex tasks require multiple persons to cooperate mentally and physically. For example, in the Space Shuttle STS-49 mission \cite{NASA,Orloff1992}, NASA originally planned to have only one astronaut to capture the slow-rotating satellite IntelSat but failed to accomplish the task. The task failed again the second day with two astronauts working together. The task was finally accomplished on the fifth day by three astronauts and one robot (the Canadarm) worked together. The mission set several records including maximum number of astronauts in space walk and longest hours in a single spacewalk \cite{NASA}. The main reason such task requires multiple astronauts is that the size of the IntelSat is overwhelming for a single astronaut. We can easily imagine that a single robot is facing the same challenge when handling oversized objects. A multi-robot system (MRS) can achieve the goal but deploying such a system requires more advanced coordination and control strategies. 

Oversized object transportation is a typical task that is usually composed with smaller subtasks which can be assigned to multiple robots simultaneously \cite{tuci2018cooperative}. We assume that the robots are physically attached to the object, and transport is achieved by either pushing or pulling (or both) the object. Decentralized architecture is the natural way to control a MRS in such a task as it is more flexible and more scalable \cite{ismail2018survey}. To solve such a problem, on the one hand, an individual robot has to guarantee accomplishment of its own assignment. On the other hand, the robots have to cooperate with each other to achieve the shared high-level goal. However, the variation of the system (e.g. number of the robots, moving obstacles, unpredictable perturbations, etc.) can still bring challenges toward management of the MRS. A leader/follower architecture is very popular in the past for it can plan and control on top of the leader robot's well-studied dynamics model \cite{kosuge1996decentralized,wang2016kinematic}. Besides of that, researchers has proved that MRS with identical controllers equally distributed on individuals performs the task well \cite{farivarnejad2016decentralized}. However, treating all the members equally can results in a more complicated dynamics model of the integrated system.

Designing a good controller for a robot could be extremely challenging due to the complicated dynamics of the task. However, the performance of a controller can be easily evaluated through the more obvious success conditions. Instead of modeling dynamics of the system, reinforcement learning (RL) algorithms model the reward mechanisms which directly tells the goodness of a state that is the consequence of the controller's outputs. Therefore, we can optimize the controller with the guide of the reward function to achieve the goal without knowing the dynamics model at all \cite{sutton2018reinforcement}. The rapidly developing deep learning technologies further boost the RL into deep reinforcement learning (DRL) which enables handling more representative data (e.g. image) and thus has been applied to many complex control problems \cite{mnih2015human}. DRL is also welcome by the researchers aiming at solving cooperative object transportation tasks with MRS \cite{gupta2017cooperative,manko2018adaptive,hernandez2019agent}.

In this paper, we present an MRS controlled by distributed DRL controllers performing cooperative object transportation task. We modified original DQN algorithm \cite{mnih2015human} to form two algorithms to train controllers in an MRS with homogeneous or heterogeneous robots. Specifically, we instantiate the MRS with two homogeneous mobile robots with each robot driven by a differential driving mechanism. The task is transporting a long rod from inside of a room to the outside of it. The most challenging part is the robots need to transport the rod through a narrow doorway. We employ DQN controllers to regulate behaviors of the robots by estimating the total reward of available actions at any given state then take the action with highest value. Due to the DQN controller's fundamental mechanism of estimating values of actions given states, we propose to use absolute error of estimated state-action values between two robots to quantitatively measure how well they cooperate with each other. To our best knowledge, this is the first implementation of DRL algorithm in the similar context. The highlights of this research are:
\begin{itemize}
    \item Controllers are distributed on individual robots. No command center involved;
    \item No dynamics modeling, no path planning. Controllers make decisions directly from the sensing data (end-to-end);
    \item The progress of gradually unified value estimations is observed through the newly proposed cooperation metric;
\end{itemize}
In the next section, we will review some related researches; we introduce our research methodology in section \ref{sec:methodology}; the experiment results and analyses will be described in section \ref{sec:results_and_analyses}. Stay tuned . . .

\section{Related Work}
\label{sec:related work}
More researchers began to be interested in solving the problem of cooperative object transportation using MRS from mid 90's \cite{kube1993collective,wang1994cooperating,sen1994learning,brown1995pusher,kosuge1996decentralized}. Although the settings could be varied a lot from each others, there are three major strategies to configure such a problem: 1) pushing-only strategy; 2) grasping strategy; 3) caging strategy \cite{tuci2018cooperative}. Our research adopt the second strategy as it keeps the complexity of the task that the robots have to coordinate with each others using different actions(pushing or pulling), while the robots are focusing on the transportation without considering the spatial configuration. While centralized control schemes are rarely reported, There exist research using such an organization \cite{hichri2016cooperative}. Nevertheless, majority of the researchers adopt the distributed architectures, thus the following literatures are default with decentralized control schemes.

When solving MRS based cooperative object transportation problem with grasping strategy, leader/follower configuration is very popular \cite{kosuge1996decentralized,machado2016multi,wang2016kinematic,bechlioulis2018collaborative,lin2018interval}. In general, a leader robot is responsible for initiating and directing the transportation, while the follower robots coordinate their actions with respect to the leader's guidance. It is true that the leader/follower architecture saves the cost of computation and communication, but this architecture sacrifices some flexibility of the follower robots. Under some complicated situations, involving a leader robot with superior capabilities does not make the problem easier. 

Researchers investigated MRS with every individual playing the same role, such that the MRS can be more flexible in their tasks. A two-stage motion planning strategy was proposed in \cite{alonso2017multi,alonso2019distributed} to help the MRS safely transport an object around dynamic obstacles. A research scenario that was closely analogous to ours was proposed in \cite{ponce2016cooperative}. The researchers were able to control two omnidirectional mobile robots to transport an object through an narrow opening with decentralized sliding mode controllers. In this research, predefined trajectories and dynamics model of the MRS needs to be established beforehand. An MRS controlled with decentralized sliding mode controllers was reported in \cite{farivarnejad2016decentralized}, which was able to transport arbitrary shaped objects without predefined trajectories shaped. However, dynamics model of the individual robot is still needed. Research of a decentralized adaptive control strategy were reported in \cite{verginis2019robust}, which enabled cooperative object transportation with two robotic arms. These researchers also investigated model predictive control (MPC) on this task \cite{nikou2017nonlinear}. Although adaptive control compensated uncertainties and MPC deal with unsolvable optimalities, the dynamics model and path planning routine cannot be saved. Our colleagues proposed a solution with fuzzy logic system which saves the complication of dynamics modelling and path planning \cite{sun2020intelligent}. However, the control strategy is only validated with point kinematics and point-mass dynamics. 

Due to the complicated dynamics in the cooperative object transportation task, researchers began to seek help from DRL. The most related research was introduced in \cite{manko2018adaptive} that two robots transporting an object through a narrow opening managed by a DQN style algorithm \cite{watkins1992q}. However, this research relied on a path planning algorithm and the function of the DQN algorithm was only to adjust the robots whenever pre-planned trajectory is not accessible. In \cite{fernandez2015learning}, a multi-agent reinforcement learning algorithm was proposed to deal with hose transportation problem. The proposed algorithm was based on the original Q-learning, thus was not capable of continuous state inputs. A showcase of multiple robots carrying a long rod while preventing it to fall was demonstrated in \cite{gupta2017cooperative}. This research was highlighted with controlling the MRS with taking continuous input data to resolve continuous output signals. Although the algorithm was more generalized, the goal and constraints were largely different from out study.

\section{Methodology}
\label{sec:methodology}
\subsection{System Configuration}
\begin{figure}[h]
    \centering
    \includegraphics[width=0.8\textwidth]{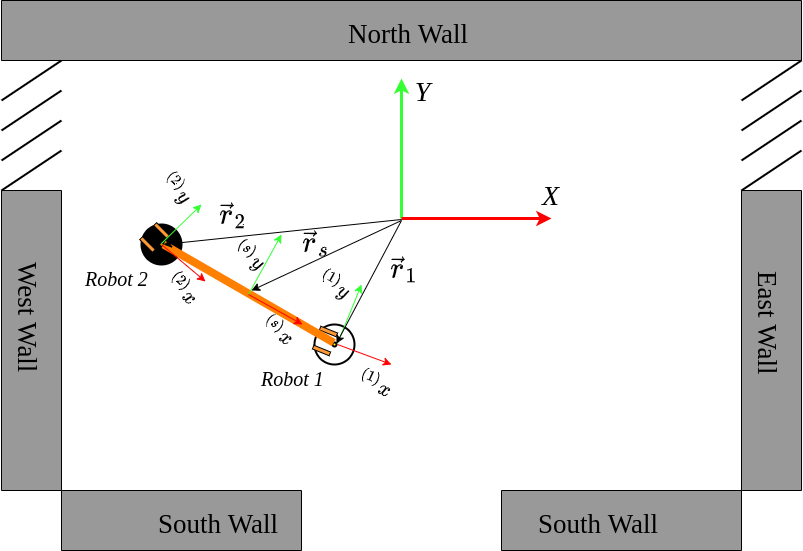}
    \caption{Task environment description}
    \label{fig:env_desc}
\end{figure}

The problem originates from a scenario that two persons carrying a big piece of furniture out of the room through a narrow door. We simplify and model the furniture transportation as two mobile robots linked by a solid rod moving out of a walled cell with an opening. We define the cell in a squared shape with dimensions of $10\times10$. A global coordinate system $\{X,Y,Z\}$ is fixed to the center of the room, where $X$ axis is pointing to the east, $Y$ axis is pointing to the north and $Z$ axis is determined by the right-hand rule. The opening is located on the south wall which is a doorway with width, $w=2$ and depth, $d=1$. The rod has length $l=w$. After attached to the robots with radius $r=0.25$, the rod cannot be transported out when parallel to the room opening. We restrict the whole system with two linear degrees of freedom (DOF) and one rotational DOF all in the plane of $XY$. Three body reference frames: $\{^{(1)}x,^{(1)}y,^{(1)}z\}$, $\{^{(2)}x,^{(2)}y,^{(2)}z\}$ and $\{^{(s)}x,^{(s)}y,^{(s)}z\}$ are attached to \textit{robot 1}, \textit{robot 2} and the rod, respectively. The $^{(1)}x$ and $^{(2)}x$ are set to point toward the head of the robots, while $^{(s)}x$ is set along the line between body frames of the robots pointing towards \textit{robot 1}'s origin. $^{(i)}z$ axes are perpendicular to the $XY$ plane with same direction as $Z$ axis point to. Hence, $^{(i)}y$ axes can be determined by $^{(i)}x$, $^{(i)}z$ and right hand rule. Coordinates of body frames origins under the global frame define positions of the robots ($\Vec{r}_1$, $\Vec{r}_2$) and the rod ($\Vec{r}_s$). Our settings can be illustrated by Fig. \ref{fig:env_desc}.

\subsection{Deep Reinforcement Learning Context}
The rod transportation task can be viewed as a series of events. The task starts at time step $t=0$ and ends at time step $t=T$. Everything happens in between this period consists an episode. At any time step $t$, we can describe a robot in the state of $^{(i)}\mathbf{s}_t$, where $i$ is the index of the robot. Assume every robot in this system can perfectly sense poses and rotations of the rod and all the robots (include itself and teammates), we can define the state of \textit{robot i} as: $^{(i)}\mathbf{s}_t=[^{(i)}\Vec{r}_t,^{(i)}\dot{\Vec{r}}_t,^{(s)}\Vec{r}_t,^{(s)}\dot{\Vec{r}}_t,^{(j)}\Vec{r}_t,^{(j)}\dot{\Vec{r}}_t]$, where $\dot{\Vec{r}}$'s are velocity vectors of the robots and the rod. Each robot can take an action $^{(i)}\mathbf{a}_t\in\{\textit{forward left, forward right, backward left, backward right}\}$ at any time step $t$. Then, the system will transfer to the next step $t+1$, and the state of \textit{robot i} can be obtained as: $^{(i)}\mathbf{s}_{t+1}=[^{(i)}\Vec{r}_{t+1},^{(i)}\dot{\Vec{r}}_{t+1},^{(s)}\Vec{r}_{t+1},^{(s)}\dot{\Vec{r}}_{t+1},^{(j)}\Vec{r}_{t+1},^{(j)}\dot{\Vec{r}}_{t+1}]$. Together with the new state, the \textit{robot i} receives a reward $^{(i)}\mathbf{r_{t+1}}$ as defined in Eq. \ref{eq:sparse_rew} which is the key to guarantee the robots can cooperate in this task. An individual will not receive any reward until both of them successfully escaped the room.
\begin{equation}
    ^{(i)}\mathbf{r}(^{(i)}\mathbf{s},^{(i)}\mathbf{a}) =
    \begin{cases}
        1 & \text{if entire body of the rod is out of the room}\\
        0 & \text{otherwise}
    \end{cases} 
    \label{eq:sparse_rew}
\end{equation}
However, rewards generated by such function will be too sparse that the learning process could be largely slow down due to this effect. In practice we extend Eq. \ref{eq:sparse_rew} to the form as seen in Eq. \ref{eq:dense_rew} to guarantee non-zero reward can be received at any time step.
\begin{equation}
    ^{(i)}\mathbf{r}(^{(i)}\mathbf{s},^{(i)}\mathbf{a}) =
    \begin{cases}
        400 & \text{if entire body of the rod is out of the room}\\
        -100 & \text{if any robot hit wall} \\
        -0.1 & \text{otherwise}
    \end{cases} 
    \label{eq:dense_rew}
\end{equation}
The dense reward function in Eq. \ref{eq:dense_rew} is inspired by \textit{LunarLander} environment from OpenAI Gym \cite{brockman2016openai}. Instead of giving negative reward according to scale of control signals, we punish the robots from the perspective of time. Since we limit the horizon of an episode to be 1000 time steps, it is reasonable to give a large negative reward for the event of hitting the wall 1000 times larger than the routine time cost. In our case, if punishment for wall hitting larger than -100 (e.g. -1), the robots are possible to stuck at a local optimal which will lead them hitting the wall directly. Because they are more likely to receive smaller negative total reward for hitting the wall compare to struggling too long in the room but accumulate larger negative reward in the end. 

In an episode, the total reward a robot receives by taking an action $\mathbf{a}_t$ at state $\mathbf{s}_t$ can be defined as
\begin{equation}
    \mathbf{R}_t = \sum_{k=0}^{T-t}\gamma^k\textbf{r}_{t+k+1}(\mathbf{s}_t,\mathbf{a}_t) 
    \label{eq:total_rew}
\end{equation}
where $\gamma\in[0,1]$ is the discount rate which weighs future rewards less and less because of the nature of uncertainties. We can further describe the value of an action $\mathbf{a}_t$ taken at state $\mathbf{s}_t$ to be the expected total reward: $\mathbf{Q(\mathbf{s}_t,\mathbf{a}_t)}=\mathbb{E}[\mathbf{R}_t]$. We can call this value of state-action pair as \textbf{Q value}. Assume the behavior of a robot is determined by a control policy $\pi(\mathbf{a}|\mathbf{s})$, then the Q value of current step can be represented by the Q value of next time step according to the Bellman Expectation Equation as shown in Eq. \ref{eq:bellman_expect} \cite{sutton2018reinforcement}. 
\begin{equation}
    \mathbf{Q}^{\pi}(\mathbf{s}_t,\mathbf{a}_t) = \mathbf{r}_{t+1}+\gamma\sum_{\mathbf{s}_{t+1}\in\mathcal{S}}\mathcal{P}(\mathbf{s}_{t+1}|\mathbf{s}_t,\mathbf{a}_t)\sum_{\mathbf{a}_{t+1}\in\mathcal{A}}\pi(\mathbf{a}_{t+1}|\mathbf{s}_{t+1})\mathbf{Q}^{\pi}(\mathbf{s}_{t+1},\mathbf{a}_{t+1})
    \label{eq:bellman_expect}
\end{equation}
where $\mathcal{P}$ is the dynamics model (here we use a probability model) which governs the transition between two consecutive time steps. Typically, we need to model the dynamics, $\mathcal{P}$, such that we can make plans for the robots then control it to stick to the plan. However, obtaining the dynamics model becomes more challenging as the task getting more complex. Reinforcement learning (RL) methods seek to solve the problem bypassing the dynamics model to only focus on optimizing the control policy through trial-and-error style interactions. The interactions has one objective that is maximizing the expected total reward at any given state, thus the dynamics model can be safely ignored. In the MRS cooperative object transportation problem, dynamics of the system is hard to be modeled. Therefore, we propose to use RL method to solve such a problem.

\subsection{Multi-robot Deep Q-network}
To success in this task, a robot has to employ an optimal control policy, $\pi^*$ that maximizes the expected total reward at a given state, $\mathbf{s}_t$ by taking an optimal action, $\mathbf{a}_t$ governed by $\pi^*$. Then Eq. \ref{eq:bellman_expect} becomes Eq. \ref{eq:bellman_optimal} according to Bellman Optimality Equation.
\begin{equation}
    \mathbf{Q}^{\pi^*}(\mathbf{s}_t,\mathbf{a}_t) = \mathbf{r}_{t+1}+\gamma\sum_{\mathbf{s}_{t+1}\in\mathcal{S}}\mathcal{P}(\mathbf{s}_{t+1}|\mathbf{s}_t,\mathbf{a}_t)\max_{\mathbf{a}_{t+1}}\mathbf{Q}^{\pi^*}(\mathbf{s}_{t+1},\mathbf{a}_{t+1})
    \label{eq:bellman_optimal}
\end{equation}
Q-learning provides a straightforward way to iteratively optimize tabularized Q values without considering dynamics model, $\mathcal{P}$ \cite{watkins1992q}. The Q values can be updated through Eq. \ref{eq:ql} that
\begin{equation}
\mathbf{Q}(\mathbf{s}_t,\mathbf{a}_t) = \mathbf{Q}(\mathbf{s}_t,\mathbf{a}_t) + \alpha(\mathbf{r}_{t+1}+\gamma\max_{\mathbf{a}_{t+1}}\mathbf{Q}(\mathbf{s}_{t+1},\mathbf{a}_{t+1}) - \mathbf{Q}(\mathbf{s}_t,\mathbf{a}_t))
    \label{eq:ql}
\end{equation}
where $\alpha$ is the learning rate. Given the optimized Q values, a greedy policy $\mathbf{a}=\argmax_{\mathbf{a}}{\mathbf{Q}(\mathbf{s},\mathbf{a})}$ usually works well to achieve the goal. Q-learning algorithm successfully gets rid of the restriction of dynamics model, but is suffered from tabularized states and actions. From previous section, our robot's state space and action space has 18 and 4 dimensions, respectively. Discretize such state and action space may result in a giant table that requires forever to be converged. Hence, we can introduce a function approximator to take continuous states into account and serve the same role as the Q table in Q-learning algorithm. A popular type of function approximator is the neural networks (NN), and this is how Deep Q-network succeeded in the control tasks with image inputs \cite{mnih2015human}. We adopt NN with trainable weights, $\boldsymbol{\theta}$ to approximate Q function as the Q-net: $\mathbf{Q}(\mathbf{s},\mathbf{a};\boldsymbol{\theta})$. The Q-net needs to approximate the expected total reward as can be computed in Eq. \ref{eq:total_rew}. So, we need define a loss function which can tell the difference between the current Q values and the expected total rewards as Eq. \ref{eq:q_loss} shows.
\begin{equation}
    \mathcal{L}(\boldsymbol{\theta})=[\mathbf{r}_{t+1}+\gamma\max_{a_{t+1}}\mathbf{Q}(\mathbf{s}_{t+1},\mathbf{a}_{t+1};\boldsymbol{\theta})-\mathbf{Q}(\mathbf{s}_t,\mathbf{a}_t;\boldsymbol{\theta})]^2
    \label{eq:q_loss}
\end{equation}
Differentiating Eq. \ref{eq:q_loss} with respect to $\boldsymbol{\theta}$, we obtain the gradient $\nabla_{\boldsymbol{\theta}}\mathcal{L}$. Hence we can use stochastic gradient decent to update $\boldsymbol{\theta}$ and optimize the loss function.

The robots in this task are set to be homogeneous, thus only one DQN controller needs to be trained with integration of all the robots' experience. The individuals in the MRS can actually accelerate the training by collecting more data in every time step. A little modification on the original DQN algorithm leads to our multi-robot DQN algorithm for homogeneous MRS as shown in Alg. \ref{alg:homo_dqn}
\begin{algorithm}[h]
    \caption{Homogeneous MRS DQN}
    \label{alg:homo_dqn}
    \begin{algorithmic}
        \REQUIRE Initialize replay memory $\mathcal{D}$ \\
        \REQUIRE Initialize active Q-net with random weights: $\boldsymbol{\theta}$ 
        \REQUIRE Initialize target Q-net with identical weights as in the active Q-net: $\boldsymbol{\bar{\theta}} = \boldsymbol{\theta}$
        \FOR{$episode=1$ \TO $M$ } 
        \STATE {Initialize pose of the MRS randomly \\
            Perform $\epsilon$ decay
            \FOR{$step=1$ \TO $T$ } 
                \FOR{$robot=1$ \TO $N$ }
                \STATE {
                    Select a random action$^{i}\mathbf{a}_{t}$ with probability $\epsilon$; \\
                    otherwise, $^{(i)}\mathbf{a}_{t}=\argmax\mathbf{Q}(^{(i)}\mathbf{s}_{t},^{(i)}\mathbf{a}_{t};\boldsymbol{\theta})$
                }
                \ENDFOR \\
                Execute all $^{(i)}\mathbf{a}_{t}$'s, then observe next states: $^{(i)}\mathbf{s}_{t+1}$, receive rewards: $^{(i)}\mathbf{r}_{t+1}$. \\
                Store all transitions $(^{(i)}\mathbf{s}_{t},^{(i)}\mathbf{a}_{t},^{(i)}\mathbf{s}_{t+1},^{(i)}\mathbf{r}_{t+1})$ into $\mathcal{D}$ \\
                Sample random batch of transitions $(\mathbf{s}_j,\mathbf{a}_j,\mathbf{s}_{j+1},\mathbf{r}_{j+1})$ from $\mathcal{D}$ \\
                Set target $y_{j}=\mathbf{r}_{j+1}$ if episode terminates at $j+1$ \\
                otherwise, $y_{j}=\mathbf{r}_{t+1}+\gamma\mathbf{Q}(\mathbf{s}_{j+1},\argmax_{\mathbf{a}}\mathbf{Q}(\mathbf{s}_{j+1},\mathbf{a};\boldsymbol{\theta});\bar{\boldsymbol{\theta}})$ \\
                Perform gradient decent on $\boldsymbol{\theta}$ \\
                Every $C$ steps, reset $\boldsymbol{\bar{\theta}} = \boldsymbol{\theta}$
            \ENDFOR
        }       
        \ENDFOR
    \end{algorithmic}
\end{algorithm}
Although the number of robots in this research is limited to two, this algorithm can be scalable to more robots as needed. As opposed to the original DQN algorithm, we implement Double DQN (DDQN) trick to compute target Q values so that the algorithm can suppress the over-optimistic estimations toward the target Q values \cite{van2016deep}. We also extend Alg. \ref{alg:homo_dqn} to a heterogeneous version (Alg. \ref{alg:hete_dqn}) so that an individual in an MRS will be trained by self collected data only. As homogeneous MRS is just a special case of heterogeneous MRS, Alg. \ref{alg:hete_dqn} can adaptive to the homogeneous MRS without any problem. 

\subsection{Cooperation Metrics}
\label{subsec:cooperation_metrics}
As researchers lack of tools to quantitatively assess the cooperation among the individuals in an MRS, we propose to introduce the mean absolute error (MAE) of Q-values between any two robots to be the standard. The nature of the Q-value is the expected total reward as being illustrated by Eq. \ref{eq:bellman_expect}. At time step $t$, the expected total reward of robot $i$ can be represented by $\max_\mathbf{a}(^{(i)}\mathbf{Q}(^{(i)}\mathbf{s}_t,^{(i)}\mathbf{a}_t))$. The absolute error of Q-values at same moment $t$ between robot $i$ and robot $j$, $\mathbf{e}_t = |^{(i)}\mathbf{Q}_t - ^{(j)}\mathbf{Q}_t|$, represents how different these two robots evaluate same moment. Using the MAE of Q-values, we introduce the novel metric $\Delta Q$ to assess cooperation between any couple of robots (Eq. \ref{eq:q_mae}).
\begin{equation}
    ^{(ij)}\Delta Q = \frac{1}{T}\sum_{t=0}^T|^{(i)}\mathbf{Q}_t - ^{(j)}\mathbf{Q}_t|
    \label{eq:q_mae}
\end{equation}

\subsection{Experiment Configurations}
Design of the two-robot system is as Fig. \ref{fig:logger_model} illustrated in Appendix \ref{sec:robot_dimensions}. For each individual robot, five rigid parts with basic geometries are included (chassis, left wheel, right wheel, caster wheel, hat). The rod are attached to the robots through their hats which can be freely rotated with respect to their chassis. 

\begin{figure}[h]
    \centering
    \includegraphics[width=0.4\textwidth]{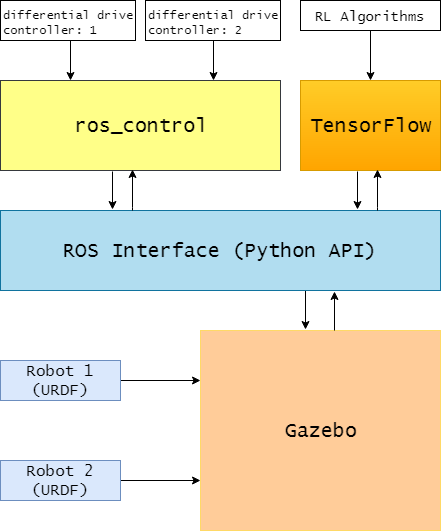}
    \caption{Simulation layout}
    \label{fig:sim_conf}
\end{figure}

The multibody dynamics is taken care by Open Dynamics Engine (ODE) \cite{drumwright2010extending}, which is integrated in Gazebo simulation software \cite{koenig2004design}. To enable interactions between the MRS and the simulated task environment, we first design robot models and environment model, then add a Python API for MRS to interact with the environment. Lastly, we can implement out RL algorithms through Tensorflow (an end-to-end open source platform for machine learning) \cite{abadi2016tensorflow}. The architecture of the software interface can be seen in Fig. \ref{fig:sim_conf}. 
The details of system modeling, software API construction and training scripts are all open-sourced and can be found at \href{https://github.com/IRASatUC/two_loggers}{https://github.com/IRASatUC/two\_loggers}.

The DQN in this research were all constructed with two fully connected hidden layers with 256 weights in each layer. Hyper-parameters we applied in the training were listed in Table \ref{tab:hyp_params} in Appendix \ref{sec:dqn_hyper-parameters}. The trainings were performed on a desktop computer with an AMD Threadripper 1900X CPU and an Nvidia GeForce GTX 1080Ti GPU. A GPU is not necessary for such task.

\section{Results and Analyses}
\label{sec:results_and_analyses}
\subsection{Training Performance Improvement}
\begin{figure}[h]
    \centering
    \includegraphics[width=0.8\textwidth]{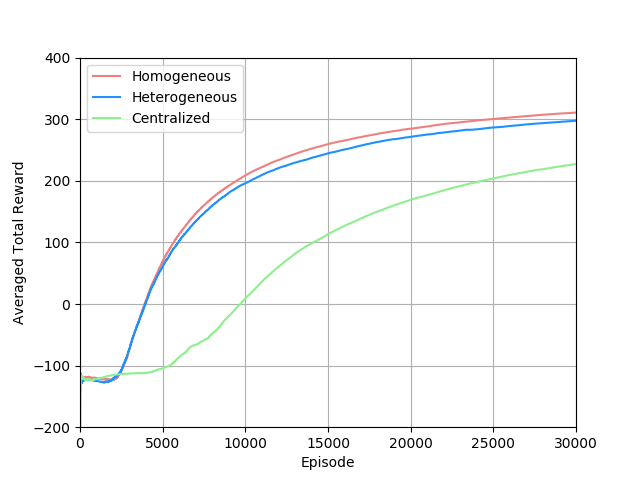}
    \caption{Averaged total reward growth along training episodes}
    \label{fig:ave_ret}
\end{figure}

We employ averaged total reward as the metric to assess performance improvement during training. Assume the MRS receives total reward $\mathbf{R}_m$ at episode $m$, then the averaged total reward at this episode can be expressed as 
\begin{equation}
    \mathbf{\bar{R}}_m = \frac{1}{m}\sum_{i=1}^m\mathbf{R}_i
    \label{eq:ave_ret}
\end{equation}
The training procedures of of using both homogeneous algorithm and heterogeneous counterparts were recorded and compared in Fig. \ref{fig:ave_ret}. In addition, we trained a centralized controller using original DQN algorithm which can be served as the baseline which is also presented in Fig. \ref{fig:ave_ret}. All these three types of training lasted 30000 episodes. We can notice that homogeneous DQN training performed slightly better than the heterogeneous one. The decentralized DQN trainings apparently outperforms centralized DQN training for they starting to success earlier but also achieved much higher averaged total reward at the 30000 episode milestone. Considering centralized DQN controller needs to handle 16 actions (compare 4 actions for decentralized setting), it is reasonable that the learning speed slowed down and the training process was less stable. The total amount of interactions(time steps) happened during homogeneous DQN training was about $5.1\times10^6$, the training of heterogeneous DQN took more than $5.8\times10^6$ interactions, while training of centralized DQN took more than $8\times10^6$ interactions to finish 30000 episodes. This result suggests that decentralized architecture is more superior than centralized counterpart. Employing homogeneous robots in an MRS is the most efficient configuration when training decentralized DQN controllers.

\subsection{DQN Controller Performance Analysis}
We evaluated the performance of the controllers by running 1000 trials with the MRS randomly initiated in the cell. The homogeneous training algorithm resulted in a MRS with 0.966 success rate, the heterogeneous training algorithm produced a MRS with 0.955 success rate, and centralized DQN gave out an answer of 0.941 success rate. We also tested performance of the DQN controllers against uncertainties. By increasingly adding Gaussian noise to the states (inputs) of the controllers, we found the success rate of both MRSs degraded gradually. Table \ref{tab:perf_noise} shows the performance drop of the MRS against the increased noise level. Considering the success rate of untrained controller (analogous to taking random actions all the time) barely reached 0.001, the DQN controllers demonstrated reasonable robustness against input uncertainties. Especially when noise level was controlled under $\mathcal{N}(0,0.1)$, the performance of decentralized DQN were not affected at all.  
\begin{table}[h]
    \centering
    \caption{Effect of state noise}
    \label{tab:perf_noise}
        \begin{tabular}{c c c c}
            \hline
            \textbf{Noise Level} & \textbf{Homogeneous} & \textbf{Heterogeneous} & \textbf{Centralized} \\
            w/o noise & 0.966 & 0.955 & 0.941 \\
            $\mathcal{N}(0,0.1)$ & 0.974 & 0.969 & 0.933 \\
            $\mathcal{N}(0,0.2)$ & 0.926 & 0.952 & 0.896 \\
            $\mathcal{N}(0,0.3)$ & 0.819 & 0.867 & 0.794 \\
            $\mathcal{N}(0,0.4)$ & 0.696 & 0.725 & 0.670 \\
            $\mathcal{N}(0,0.5)$ & 0.546 & 0.553 & 0.482 \\
            \hline
        \end{tabular}
\end{table}

The uncertainties can happen on the output end of the controllers, thus we also tested the controllers' tolerance on the randomness of control signals. The results can be seen in Table \ref{tab:perf_rand} that the performance degradation was slowly building up along the gradually increased output randomness. Even with 50\% chance that a controller will take a random action, all the controllers were still having success rate over 0.8. When output randomness was under 10\%, there was no performance drop observed in all three categories. When output randomness increased to 20\%, decentralized DQN controllers can still maintain their performance, whereas the performance of centralized controller began to degrade. 
\begin{table}[h]
    \centering
    \caption{Effect of action randomness}
    \label{tab:perf_rand}
        \begin{tabular}{c c c c}
            \hline
            \textbf{Random Level} & \textbf{Homogeneous} & \textbf{Heterogeneous} & \textbf{Centralized} \\
            w/o randomness & 0.966 & 0.955 & 0.941\\
            10\% & 0.961 & 0.969 & 0.938 \\
            20\% & 0.970 & 0.967 & 0.924 \\
            30\% & 0.952 & 0.940 & 0.925 \\
            40\% & 0.938 & 0.919 & 0.889 \\
            50\% & 0.882 & 0.884 & 0.832 \\
            \hline
        \end{tabular}
\end{table}

\subsection{Quantitative Cooperation Assessment}
As a highlight of this research, we proposed a metric, $\Delta Q$ (described in \ref{subsec:cooperation_metrics}), to assess the cooperation between two individuals in an MRS. Table. \ref{tab:q_diff} shows result from 1000 test trials with MRS randomly initiated in the cell. Under the assessment of $Delta Q$, controllers trained by homogeneous DQN algorithm demonstrated best cooperation with minimum diverse when estimating total reward at the same moment. The two controllers were having larger disagreement if trained by heterogeneous DQN algorithm. During training, we saved DQN models every $10^6$ interactions, so that changed behaviors of the robots can always be tracked by loading previously saved models. We also compared metric $\Delta Q$ at different stage of training. Both algorithms exhibited larger $\Delta Q$s when trained with $10^6$ interactions data. Both algorithms showed a trend of decreasing $\Delta Q$ along the increasing training episodes, which indicates both algorithms can shape the value systems in-between the controllers to a more unified form.
\begin{table}[h]
    \centering
    \caption{Metric of cooperation}
    \label{tab:q_diff}
        \begin{tabular}{c c c}
            \hline
            \textbf{Model version} & \textbf{Homogeneous} & \textbf{Heterogeneous} \\
            $10^6$ & $11.72\pm4.55$ & $19.39\pm9.70$ \\
            $final$ & $\mathbf{3.77\pm2.20}$ & $7.83\pm3.06$ \\
            \hline
        \end{tabular}
\end{table}

\begin{figure}
    \centering
    \begin{subfigure}[b]{\textwidth}
        \centering
        \includegraphics[width=\textwidth]{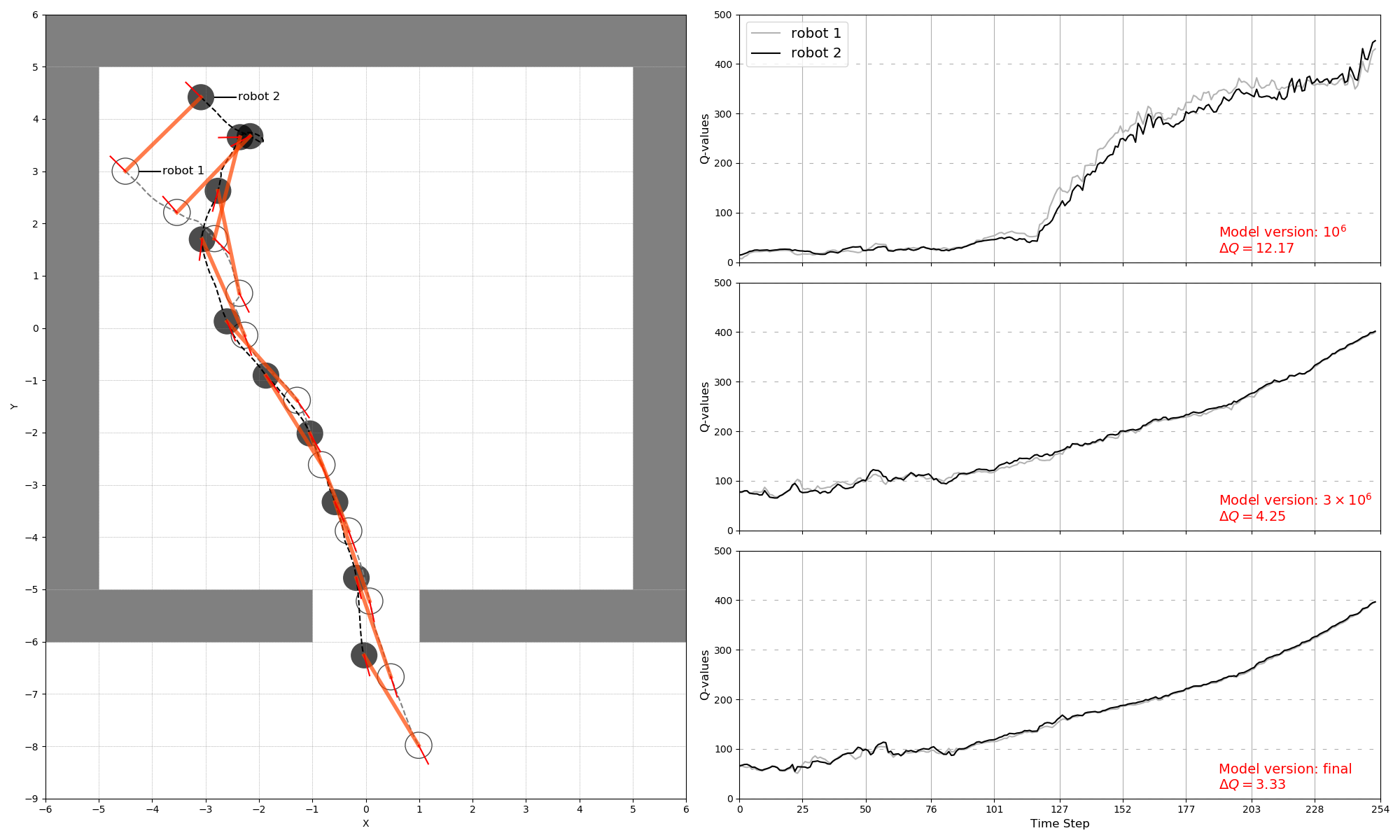}
        \caption{Homogeneous DQN algorithm trained case}
        \label{fig:homo_case}
    \end{subfigure}
    \hfill
    \begin{subfigure}[b]{\textwidth}
        \centering
        \includegraphics[width=\textwidth]{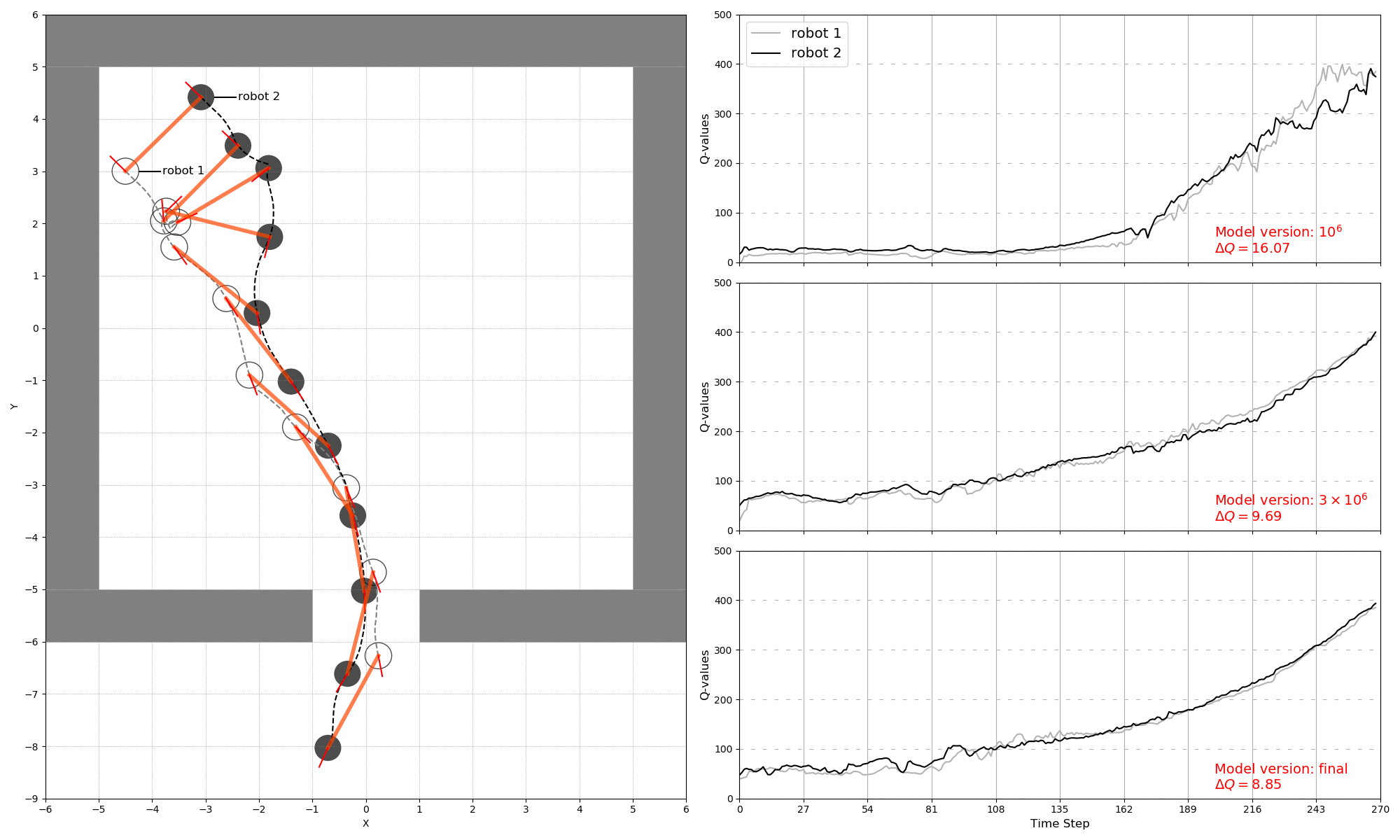}
        \caption{Heterogeneous DQN algorithm trained case}
        \label{fig:hete_case}
    \end{subfigure}
    \caption{Sampled trajectories and cooperation assessment}
    \label{fig:case_study}
\end{figure}

A case study is given here for a better understand on the cooperation metric, $\Delta Q$. We sampled two trajectories of transportation using final models trained by homogeneous DQN algorithm and heterogeneous one, respectively. We can see these two trajectories on left hand side in Fig. \ref{fig:homo_case} and \ref{fig:hete_case}. The homogeneous DQN algorithm trained controllers took more time (219 time steps) to solve the problem with \textit{robot 1} led the team. While, the heterogeneous DQN algorithm trained controllers took fewer time (205 time steps) with \textit{robot 2} led the team. According to our observations, no leader-follower pattern was formed during the training despite of algorithms. On the right hand side of Fig. \ref{fig:case_study}, we evaluate the cooperation between these two robots with different DQN models. The bottom ones indicates that homogeneous DQN algorithm trained two controllers with highly unified opinions at the same moment for Q-value difference was hardly noticed between the curves. On the heterogeneous DQN algorithm side, obvious Q-value deviations can be observed. After the trajectories being sampled, we can load different versions of DQN models to evaluate the same trajectory. Here we demonstrate how immature models value the transportation trajectories generated by the final versions of the DQN models. We pick models as being trained with $10^6$ and $3\times10^6$ interactions to represent the earlier stages of training. For both homogeneous and heterogeneous DQN algorithms, models trained with $3\times10^6$ interactions announced similar Q-values as the final models did when evaluating same situations. Besides, the cooperation metric $\Delta Q$s were relatively small. Models trained with $10^6$ interactions, however, tended to underrate the values of first 70\% trajectories. It is reasonable since the value was slowly propagate from the exit to further locations. More than that, larger $\Delta Q$ can be observed in the evaluations using models with version of $10^6$. 

\subsection{Other Findings and Discussions}
We sampled the MRS's performance from 9 different initial conditions. Metrics of \textit{time consumed}, \textit{travel distance} and $\Delta Q$ were extracted from the case-wise experiments which can be found in Table \ref{tab:case_met}. We compared performance of controllers trained by homogeneous, heterogeneous and centralized DQN algorithms. While homogeneous DQN algorithm trained controllers having significantly small gaps when evaluating states at the same moment, they were slightly outperformed heterogeneous DQN and centralized DQN trained controllers in the manners of \textit{time consumed} and \textit{distance traveled}. Trajectories of these samples were recorded as Fig. \ref{fig:trans_cases} illustrates. An obvious pattern can be observed is that the robot initially more close to the exit is more likely leading the MRS in the end. However, when the two robots were equally set against the exit, we observed that robot 1 led the MRS in most cases.

After all, DQN algorithm is not something new. Researchers proposed many more advanced DRL algorithms since the birth of DQN. A major draw back of DQN is it cannot deal with continuous action space. Deep deterministic policy gradient (DDPG) is one of the successors of DQN, which introduced a policy network to produce continuous actions \cite{lillicrap2015continuous}. Proximal policy optimization (PPO) is one of the families of policy gradient algorithms, which can optimize continuous policy through guaranteed improvement without stepping too far to avoid collapsed performance \cite{schulman2017proximal}. We have tested both algorithms in the presented task, but neither of them induced positive result so far. DDPG's convergence were extremely brittle during training, and both DDPG and PPO were having hard time to balance between exploration and exploitation. 

\section{Conclusion}
\label{sec:conclusion}
In this research, we introduced decentralized DQN controllers in an MRS to solve an object transportation task cooperatively. The controllers learned how to behave and cooperate from scratch. Given state of the MRS, the DQN controllers can output discretized control signals directly without knowing dynamics nor planning path. Two training algorithms were proposed to adapt homogeneous and heterogeneous MRS, respectively. Both algorithms were able to train well-performed DQN controllers on homogeneous robots to solve the task with high success rate. The decentralized architecture was proved to be more efficient than the centralized counterpart. The DQN controllers were proved to be robust against small to medium level uncertainties. More importantly, A novel and universal metric was proposed in this research that can quantitatively assess cooperation between robots in an MRS. With the encouraging results, we are now more confident with the potential of deep reinforcement learning (DRL) type controllers in multi-robot systems (MRSs). 

In the future, we would like to continue digging potentials of MRS with more advanced DRL algorithms, and settings. As an off-policy DRL algorithm, DQN is theoretically more data-efficient compare to the on-policy algorithms. The training process in this research still took more than $5\times10^6$ interactions or 10 days. We look forward to improve the efficiency of training with more tweaks, and this is essential to bring a DRL-type controller to a robot in the real world. The sensing data in current research is pulled out directly from the simulation software, which is not likely to be accessible in real applications. More than that, current sensing data cannot deal with constantly changing obstacles (e.g. the doorway is randomly placed). Hence, we are preparing to upgrade the current robots to be equipped with cameras and Lidars to adapt an upgraded task environment with more dynamic objects.  
\bibliography{refs.bib}
\bibliographystyle{unsrt}

\newpage
\appendices

\section{Heterogeneous MRS Training Algorithm}
Unlike the homogeneous counterpart, the training of heterogeneous MRS happens separately on each individual. Instead of sharing the data collected by all the members in the team, each robotic controller will be trained with data collected by itself.
\begin{algorithm}[h]
    \caption{Heterogeneous MRS DQN training}
    \label{alg:hete_dqn}
    \begin{algorithmic}
        \FOR{$robot=1$ \TO $N$}
        \STATE{
            Initialize replay memory $^{(i)}\mathcal{D}$ \\
            Initialize active Q-net with random weights: $^{(i)}\boldsymbol{\theta}$ \\
            Initialize target Q-net with identical weights as in the active Q-net: $^{(i)}\boldsymbol{\bar{\theta}} = ^{(i)}\boldsymbol{\theta}$
        }
        \ENDFOR
        \FOR{$episode=1$ \TO $M$ } 
        \STATE {Initialize pose of the MRS randomly \\
            Perform $\epsilon$ decay
            \FOR{$step=1$ \TO $T$ } 
                \FOR{$robot=1$ \TO $N$ }
                \STATE {
                    Select a random action$^{i}\mathbf{a}_{t}$ with probability $\epsilon$; \\
                    otherwise, $^{(i)}\mathbf{a}_{t}=\argmax\mathbf{Q}(^{(i)}\mathbf{s}_{t},^{(i)}\mathbf{a}_{t};\boldsymbol{\theta})$
                }
                \ENDFOR \\
                Execute all $^{(i)}\mathbf{a}_{t}$'s, then observe next states: $^{(i)}\mathbf{s}_{t+1}$, receive rewards: $^{(i)}\mathbf{r}_{t+1}$. \\
                \FOR{$robot=1$ \TO $N$ }
                \STATE{
                    Store transition $(^{(i)}\mathbf{s}_{t},^{(i)}\mathbf{a}_{t},^{(i)}\mathbf{s}_{t+1},^{(i)}\mathbf{r}_{t+1})$ into $^{(i)}\mathcal{D}$ \\
                    Sample random batch of transitions $(^{(i)}\mathbf{s}_j,^{(i)}\mathbf{a}_j,^{(i)}\mathbf{s}_{j+1},^{(i)}\mathbf{r}_{j+1})$ from $^{(i)}\mathcal{D}$ \\
                    Set target $^{(i)}y_{j}=^{(i)}\mathbf{r}_{j+1}$ if episode terminates at $j+1$ \\
                    otherwise, $^{(i)}y_{j}=^{(i)}\mathbf{r}_{t+1}+\gamma^{(i)}\mathbf{Q}(^{(i)}\mathbf{s}_{j+1},\argmax_{\mathbf{a}}^{(i)}\mathbf{Q}(^{(i)}\mathbf{s}_{j+1},^{(i)}\mathbf{a};^{(i)}\boldsymbol{\theta});\bar{^{(i)}\boldsymbol{\theta}})$ \\
                    Perform gradient decent on $^{(i)}\boldsymbol{\theta}$ \\
                    Every $C$ steps, reset $^{(i)}\boldsymbol{\bar{\theta}} = ^{(i)}\boldsymbol{\theta}$
                }
                \ENDFOR
            \ENDFOR
        }
        \ENDFOR
    \end{algorithmic}
\end{algorithm}

\section{Robot Dimensions}
\label{sec:robot_dimensions}
\begin{figure}[h]
    \centering
    \includegraphics[width=0.8\textwidth]{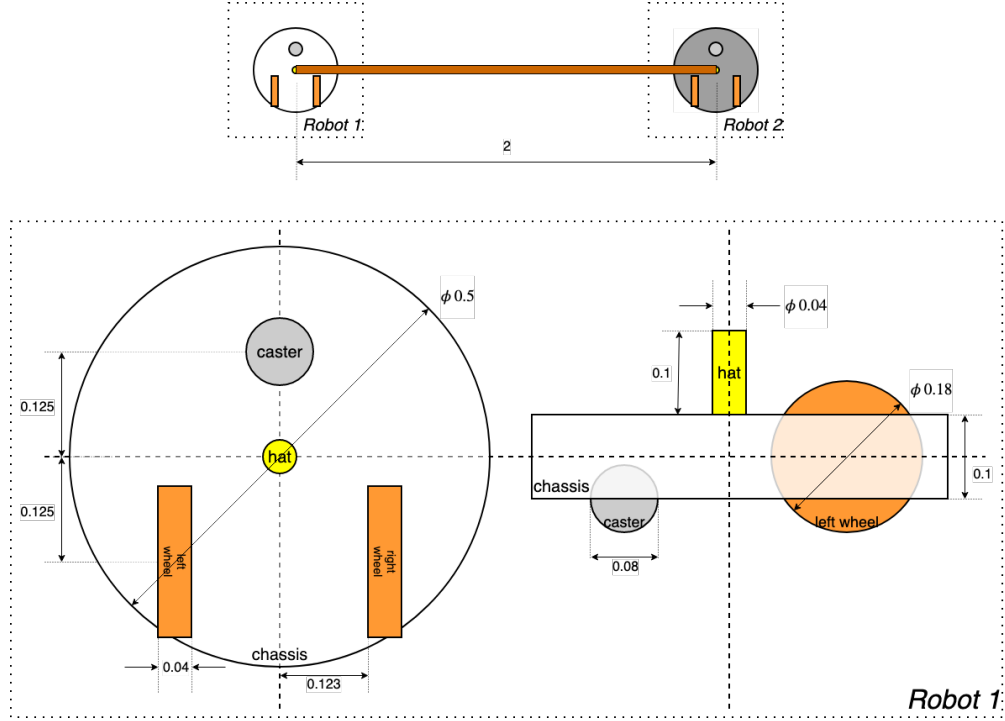}
    \caption{Dimensions of the two-robot system}
    \label{fig:logger_model}
\end{figure}

\section{DQN Hyper-parameters}
\label{sec:dqn_hyper-parameters}
Hyper-parameters used in DQN training are as follows:
\begin{table}[h]
    \centering
    \caption{DQN hyper-parameters}
    \label{tab:hyp_params}
    \begin{tabular}{m{0.2\textwidth} m{0.1\textwidth} m{0.6\textwidth}}
        \hline
        \textbf{Hyper-parameter} & \textbf{Value} & \textbf{Description} \\
        batch size & 8192 & Number of experiences sampled for one step of gradient decent optimization. \\
        replay buffer size & 10000000 & Maximum experiences can be stored in replay buffer. \\
        update frequency & 8000 & Target network will be updated to comply to the active Q-network at this frequency. \\
        discount rate & 0.99 & Used on future rewards to calculate the expected total reward. \\
        learning rate & 0.0001 & Used to control update step of the trainable weights in Q-networks. \\
        initial exploration & 1 & Initial value of action randomness. \\
        final exploration & 0.1 & action randomness after $\varepsilon$ stopped decaying. \\
        $\varepsilon$ decay period & 2000 & Number of episodes needed for $\varepsilon$ decaying from initial exploration to the final exploration \\
        warm-up episodes & 500 & Number of episodes run with completely random actions before Q-networks were optimized. \\
        \hline
    \end{tabular}
\end{table}

\section{Case Study}
\begin{figure}[h]
    \centering
    \begin{subfigure}{0.7\textwidth}
        \centering
        \includegraphics[width=0.7\linewidth, height=2.4cm]{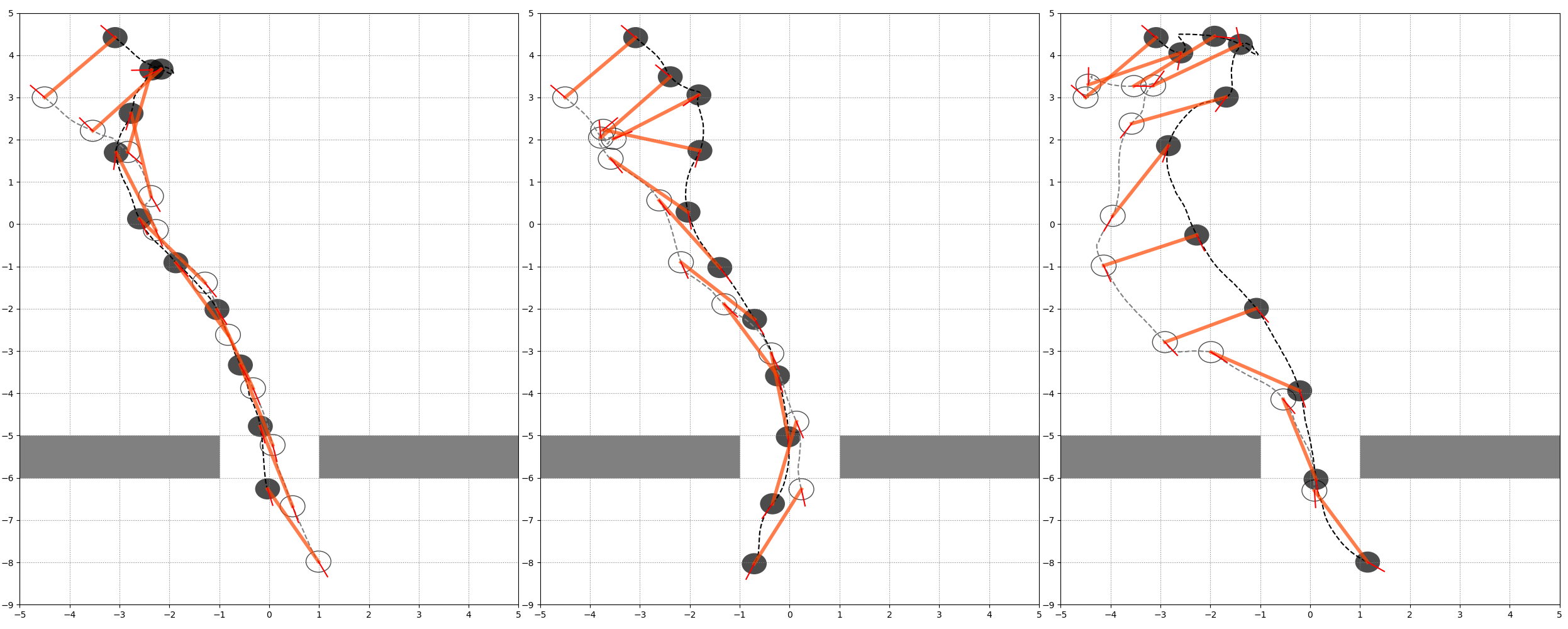} 
        \label{fig:case_1}
    \end{subfigure}
    \begin{subfigure}{0.7\textwidth}
        \centering
        \includegraphics[width=0.7\linewidth, height=2.4cm]{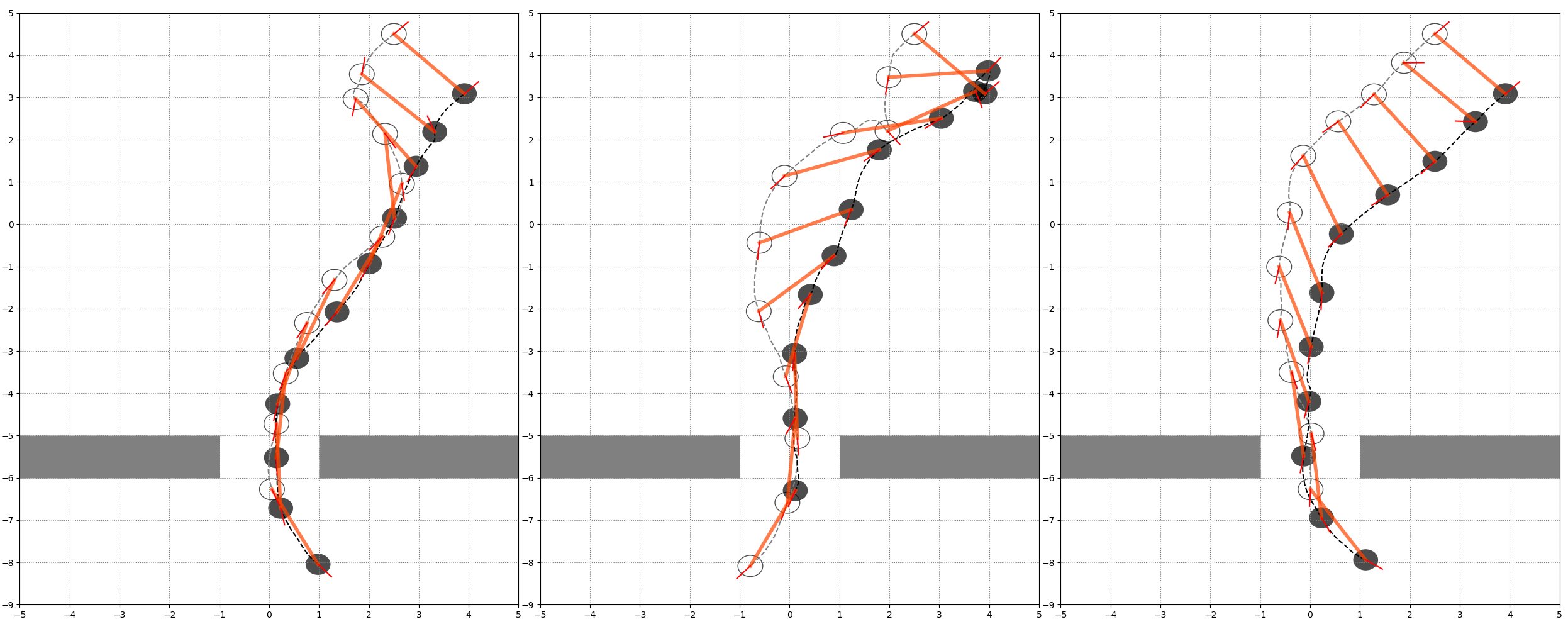}
        \label{fig:case_2}
    \end{subfigure}
    \begin{subfigure}{0.7\textwidth}
        \centering
        \includegraphics[width=0.7\linewidth, height=2.4cm]{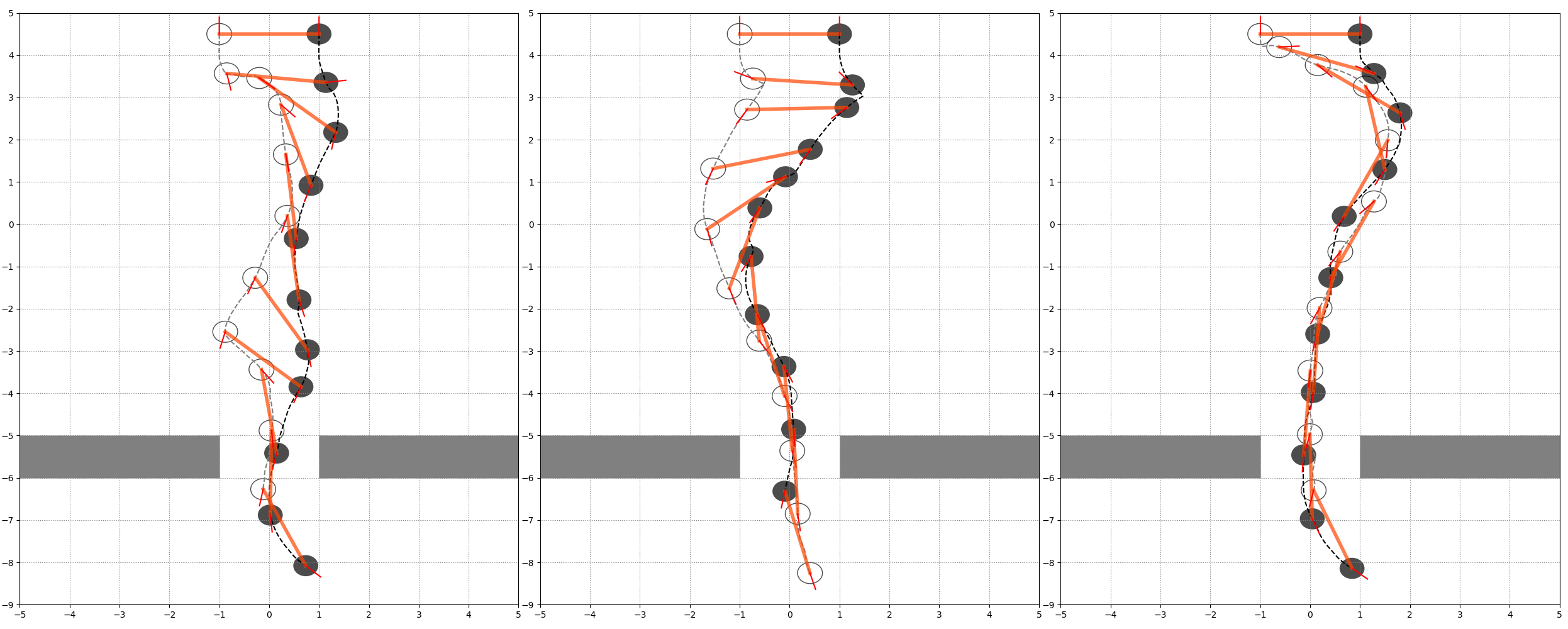}
        \label{fig:case_3}
    \end{subfigure}
    \begin{subfigure}{0.7\textwidth}
        \centering
        \includegraphics[width=0.7\linewidth, height=2.4cm]{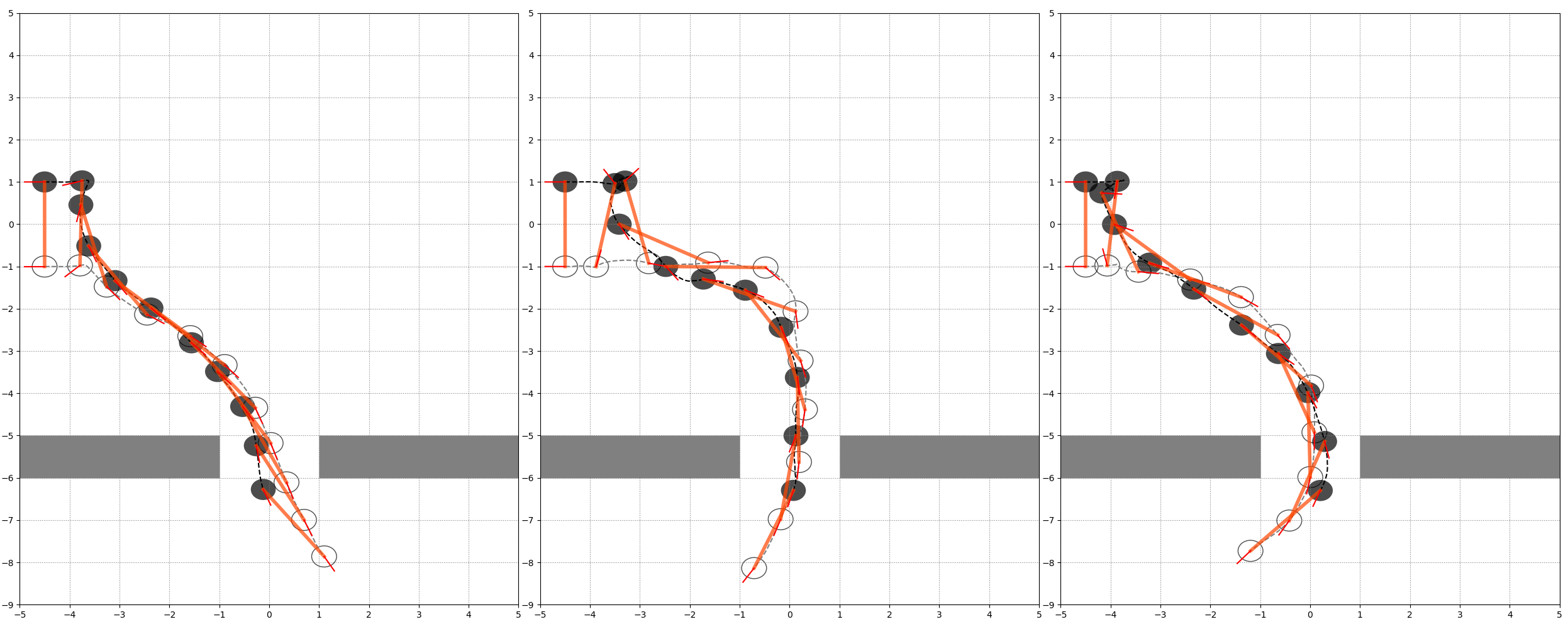}
        \label{fig:case_4}
    \end{subfigure}
    \begin{subfigure}{0.7\textwidth}
        \centering
        \includegraphics[width=0.7\linewidth, height=2.4cm]{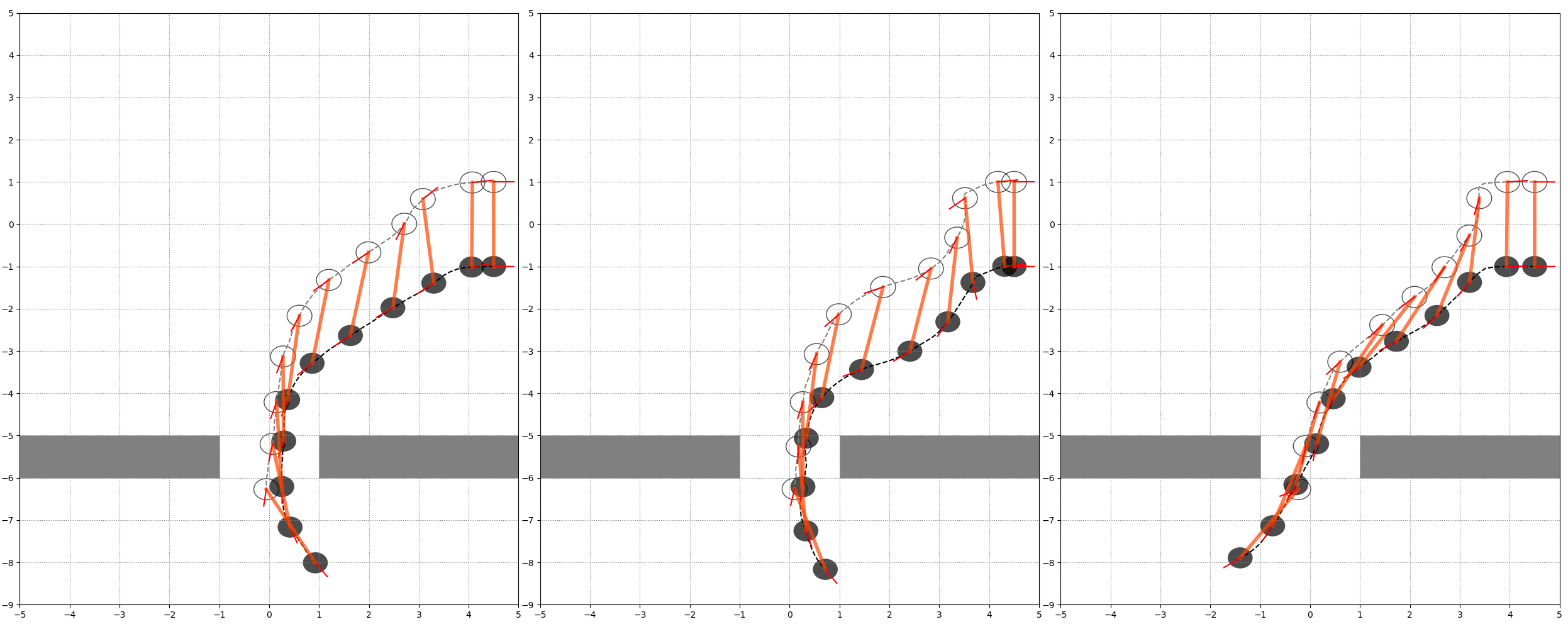}
        \label{fig:case_5}
    \end{subfigure}
    \begin{subfigure}{0.7\textwidth}
        \centering
        \includegraphics[width=0.7\linewidth, height=2.4cm]{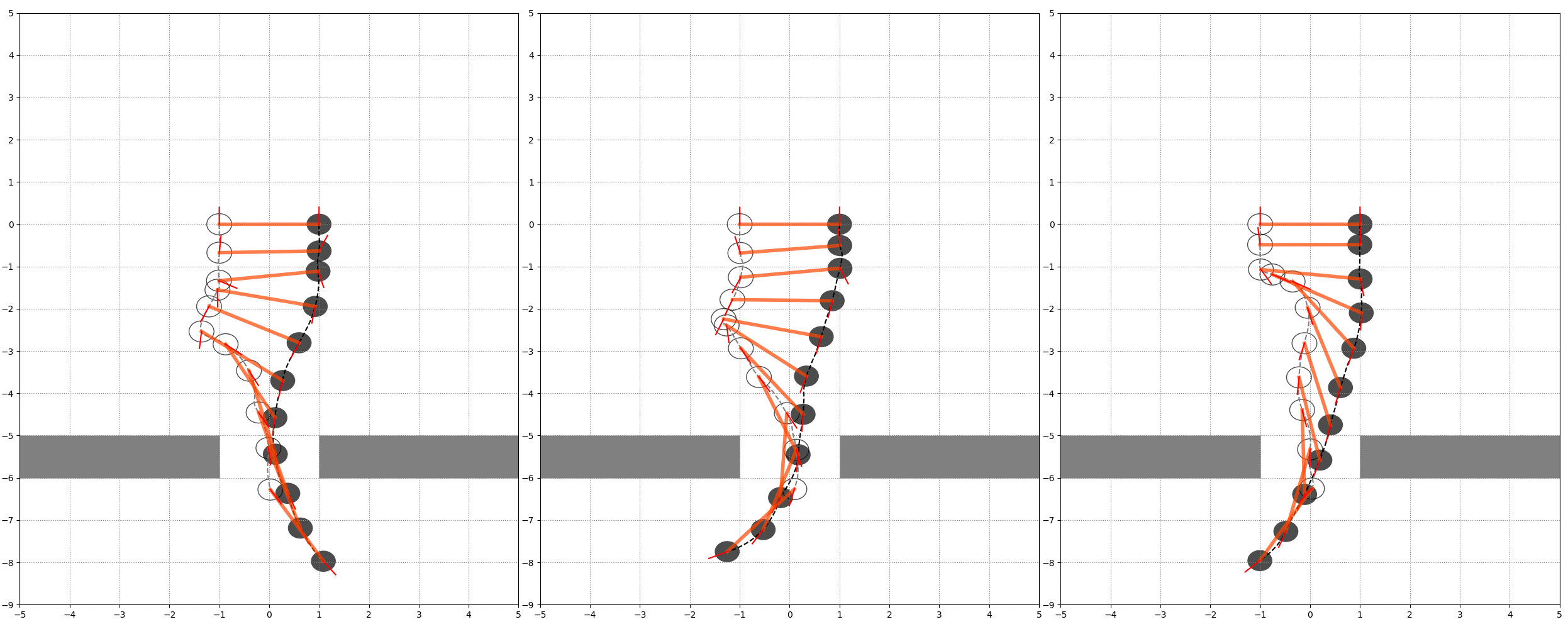}
        \label{fig:case_6}
    \end{subfigure}
    \begin{subfigure}{0.7\textwidth}
        \centering
        \includegraphics[width=0.7\linewidth, height=2.4cm]{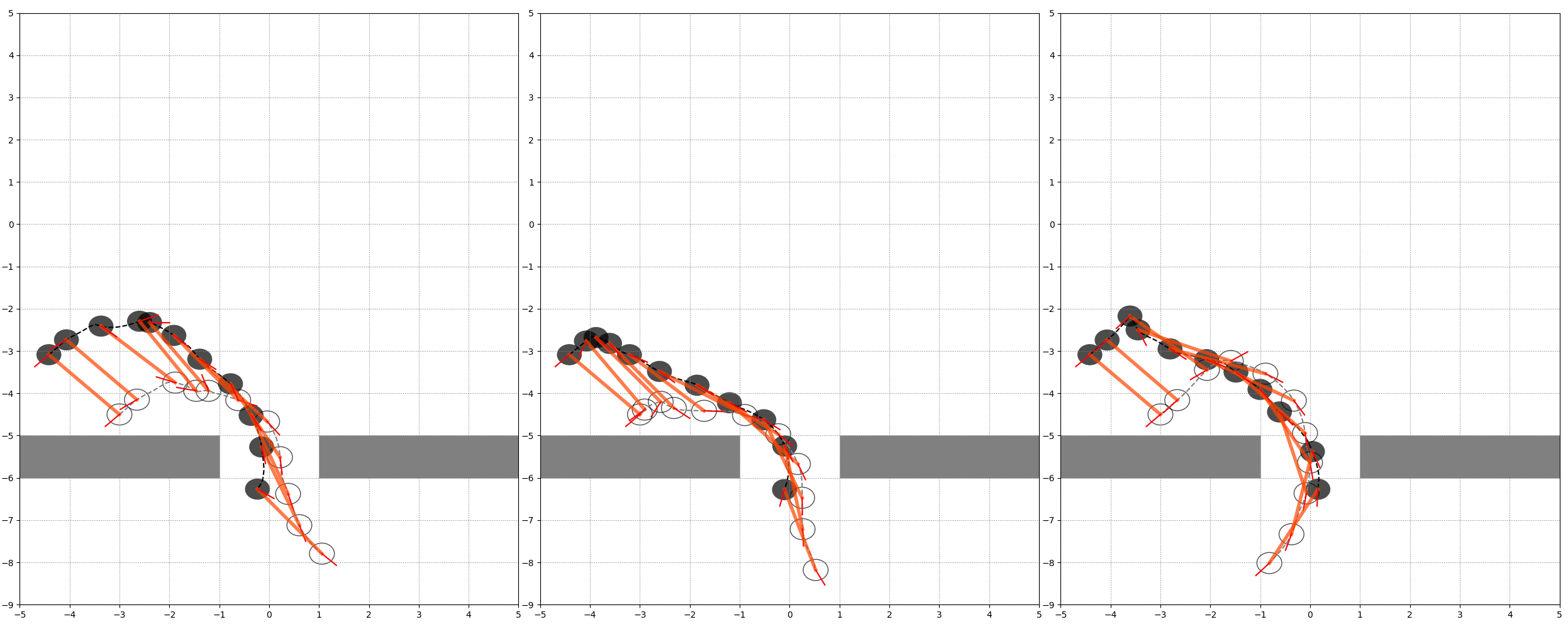}
        \label{fig:case_7}
    \end{subfigure}
    \begin{subfigure}{0.7\textwidth}
        \centering
        \includegraphics[width=0.7\linewidth, height=2.4cm]{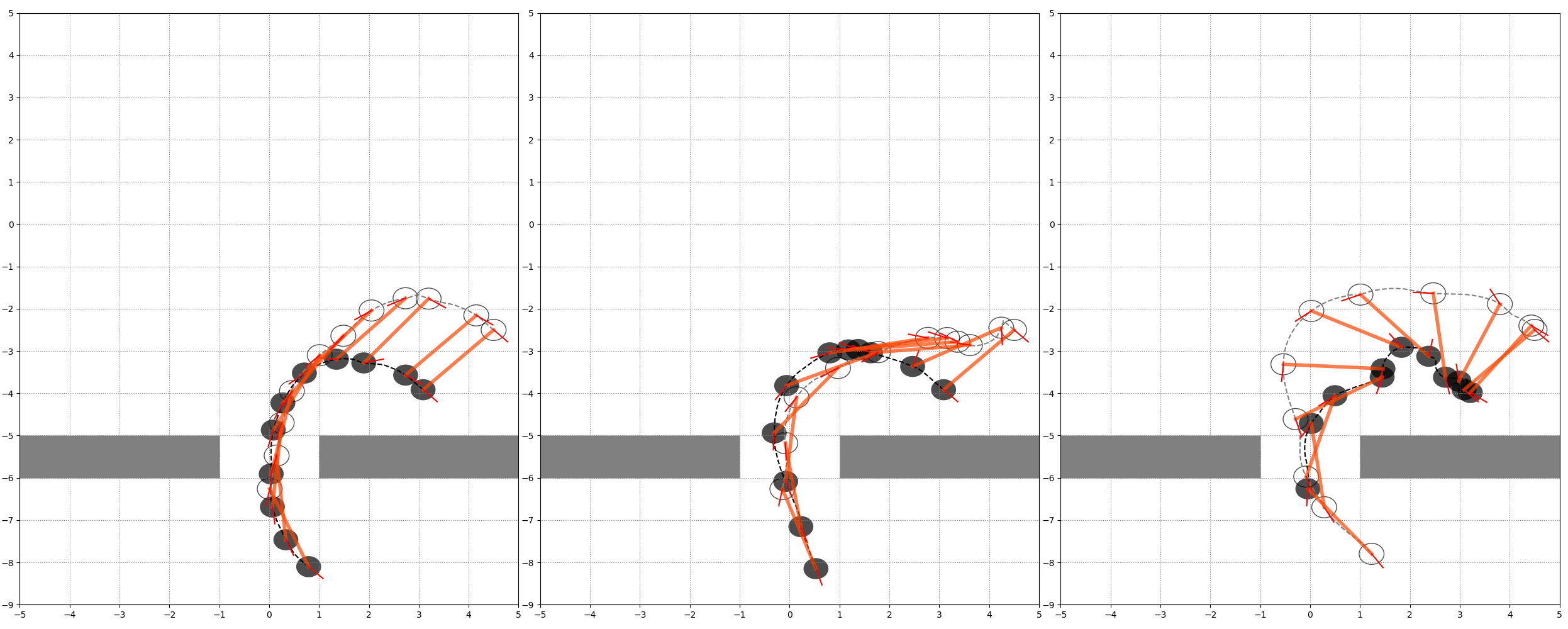}
        \label{fig:case_8}
    \end{subfigure}
    \begin{subfigure}{0.7\textwidth}
        \centering
        \includegraphics[width=0.7\linewidth, height=2.4cm]{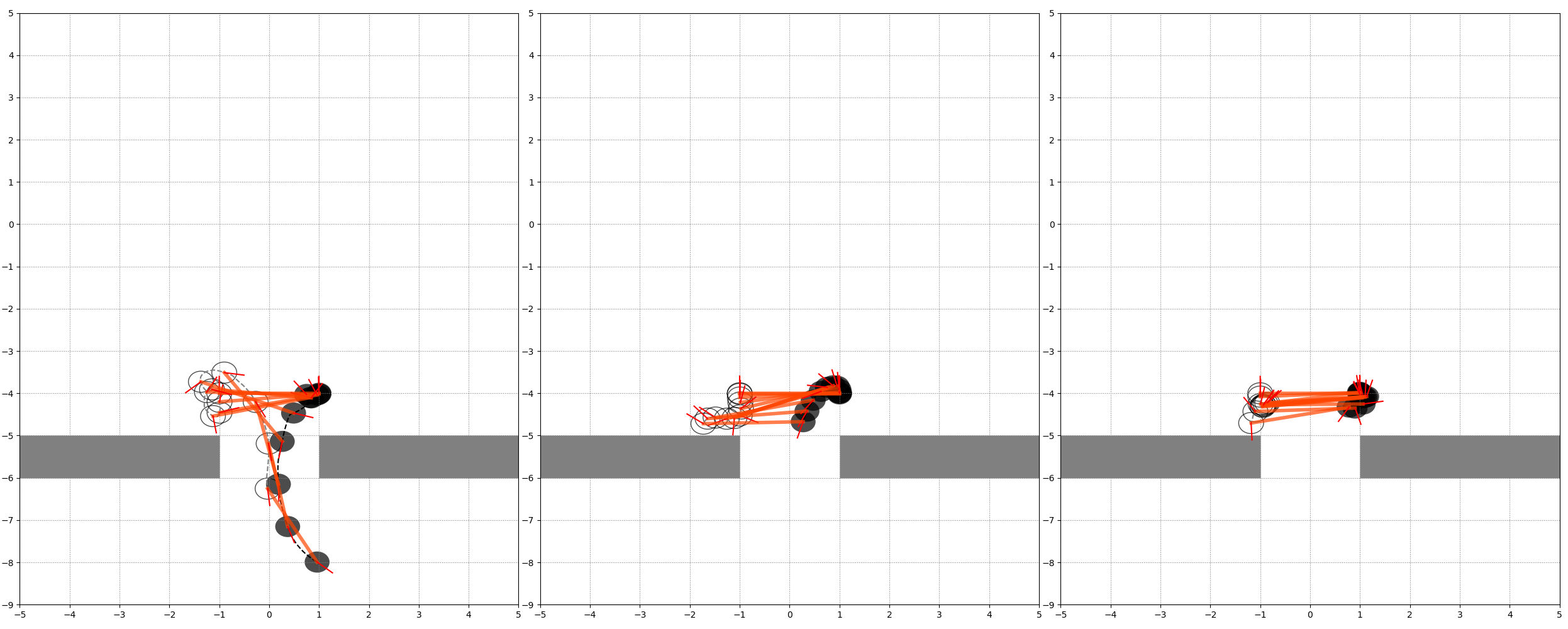}
        \label{fig:case_9}
    \end{subfigure}
    \caption{Cooperative transportation cases (left: homogeneous; middle: heterogeneous; right: centralized)}
    \label{fig:trans_cases}
\end{figure}

We sampled 9 cases of transportations with specified initial conditions. The trajectories of the transportations using different controllers can be seen in Fig. \ref{fig:trans_cases} (from left to right: homogeneous, heterogeneous, centralized). Key metrics including total travel distance, time consumed and cooperation metrics: $\Delta Q$ were recorded. Since heterogeneous and centralized DQN trained controllers failed in the last case, the assessment on these two were not applicable.
\begin{table}[h]
    \centering
    \caption{Case-wise metrics}
    \label{tab:case_met}
    \begin{tabular}{m{1.8cm} m{0.07\textwidth} m{0.07\textwidth} m{0.07\textwidth} m{0.07\textwidth} m{0.07\textwidth} m{0.07\textwidth} m{0.07\textwidth} m{0.07\textwidth}}
        \hline
        \multirow{2}{*}{\textbf{Case Index}} & \multicolumn{3}{c}{\textbf{Time Consumed}} & \multicolumn{3}{c}{\textbf{Distance Traveled}} & \multicolumn{2}{c}{$\mathbf{\Delta Q}$} \\
                                             & homo. & hete. & cent. & homo. & hete. & cent. & homo. & hete. \\
        1 & \textbf{253} & 269 & 381 & \textbf{25.62}  & 25.71 & 32.13 & \textbf{3.33} & 8.85 \\
        2 & 233 & 285 & \textbf{230} & \textbf{24.42}  & 27.28 & 24.97 & \textbf{2.54} & 6.82 \\
        3 & 265 & \textbf{252} & \textbf{252} & \textbf{25.38}  & 25.62 & 26.35 & \textbf{3.01} & 9.04 \\
        4 & \textbf{174} & 233 & 196 & \textbf{18.93}  & 22.61 & 20.38 & \textbf{3.59} & 6.41 \\
        5 & 178 & 186 & \textbf{177} & 18.92  & \textbf{18.91} & 19.12 & \textbf{2.32} & 9.86 \\
        6 & 161 & 155 & \textbf{151} & 15.24  & \textbf{15.16} & 15.22 & \textbf{3.34} & 8.68 \\
        7 & 147 & \textbf{141} & 147 & 14.19  & \textbf{12.94} & 14.89 & \textbf{5.96} & 21.90 \\
        8 & \textbf{147} & 199 & 237 & \textbf{15.30}  & 15.76 & 18.35 & \textbf{3.08} & 9.79  \\
        9 & \textbf{180} & N/A & N/A & \textbf{10.88}  & N/A & N/A & \textbf{10.11} & 15.85 \\
        \hline
    \end{tabular}
\end{table}

\end{document}